\documentclass[submission,copyright,creativecommons]{eptcs}
\providecommand{\event}{ICLP 2025} 

\providecommand{\event}{ICLP 2025} 
\usepackage{booktabs} 
\usepackage{siunitx}
\usepackage{cancel}
\usepackage{array}
\usepackage{tikz}
\usepackage{listings}
\usepackage{titlesec}
\usepackage{iftex}

\usepackage[utf8]{inputenc}
\usepackage{array, caption, tabularx, makecell, booktabs}%
\setcellgapes{3pt}
\usepackage{xcolor}
\usepackage{amsthm}
\usepackage{calc}
\usepackage{tabto}
\usepackage{lipsum}

\newtheorem{ax}{Definition}
\newcounter{examplecounter}
\newenvironment{example}{\begin{quote}%
    \refstepcounter{examplecounter}%
  \textbf{Example \arabic{examplecounter}}%
  \quad
}{%
\end{quote}%
}

\lstdefinestyle{asp}{
    basicstyle=\ttfamily\footnotesize,
    commentstyle=\color{gray},
    keywordstyle=\color{blue},
    morekeywords={:-,not,\#count,\#minimize,defeasible,classical,rank, inference, entailed, query,}
    comment=[l]{\%},        
    breaklines=true,
    alsoletter={\#},        
    frame=lines,
    captionpos=b,
    numbers=left,
    numberstyle=\tiny\color{gray}
}

\newcommand{\halmos}{$\square$} 
\newcommand{\entails}{\mid\!\sim}
\theoremstyle{definition}

\newlength{\framedinnerleftmargin}
\newlength{\framedinnertopmargin}
\newlength{\framedreversedinnerleftmargin}
{\hskip 0pt\\\hspace*{\fill}\halmos{}\end{mdframed}\vspace{1em}} 
 
{\end{spacing}\end{minipage}\end{center}} 

\titlespacing*{\paragraph}
{0pt}{3.25ex plus 1ex minus .2ex}{0pt} 
\titleformat{\paragraph}[runin]
  {\normalfont\normalsize\itshape}{\theparagraph}{1em}{}

\usepackage{xcolor}
\usetikzlibrary {positioning}
\definecolor {processblue}{cmyk}{0.96,0,0,0}
\usetikzlibrary{tikzmark,fit}
\usepackage{MnSymbol}
\usepackage{algorithm}
\usepackage{algpseudocode}
\usepackage{adjustbox}
\usepackage{graphicx}
\usepackage{mathrsfs} 
\usepackage[appendix=append]{apxproof}
\newtheoremrep{Proposition}{Proposition}

\tikzstyle{nosep}=[inner sep=0pt, outer sep=0pt]
\newcommand{\hcancel}[1]{%
    \tikz[baseline=(tocancel.base)]{
        \node[nosep] (tocancel) {#1};
        \node[nosep, yshift=.5ex]  (from) at (tocancel.south west) {};
        \node[nosep, yshift=-.6ex] (to)   at (tocancel.north east) {};
        \draw[red] (from) -- (to);
    }%
}
\ifpdf
  \usepackage{underscore}         
  \usepackage[T1]{fontenc}        
\else
  \usepackage{breakurl}           
\fi

\title{Defeasible Conditionals using Answer Set Programming}
\author{Racquel Dennison   
\institute{University of Cape Town and CAIR\\ Cape Town, South Africa}
\email{dnnrac003@myuct.ac.za}
\and
Jesse Heyninck
\institute{Open Universiteit \\ Heerlen, Limburg, the Netherlands}
\institute{University of Cape Town and CAIR\\
Cape Town, South Africa}
\email{jesse.heyninck@ou.nl}
\and
Thomas Meyer 
\institute{University of Cape Town and CAIR\\
Cape Town, South Africa}
\email{tmeyer@cair.org.za}
}
\def\titlerunning{Defeasible Conditionals using Answer Set Programming}
\def\authorrunning{Racquel Dennison, Jesse Heyninck \& Thomas Meyer}
\begin{document}
\maketitle

\begin{abstract}
Defeasible entailment is concerned with drawing plausible conclusions from incomplete information. A foundational framework for modelling defeasible entailment is the KLM framework. Introduced by Kraus, Lehmann, and Magidor, the KLM framework outlines several key properties for defeasible entailment. One of the most prominent algorithms within this framework is Rational Closure (RC). This paper presents a declarative definition for computing RC using Answer Set Programming (ASP). Our approach enables the automatic construction of the minimal ranked model from a given knowledge base and supports entailment checking for specified queries. We formally prove the correctness of our ASP encoding and conduct empirical evaluations to compare the performance of our implementation with that of existing imperative implementations, specifically the InfOCF solver. The results demonstrate that our ASP-based approach adheres to RC's theoretical foundations and offers improved computational efficiency.
\end{abstract}

\section{Introduction}
\label{section:intro}
Formal logic has played a vital role in addressing reasoning challenges in Artificial Intelligence \cite{handbook-of-knowledge-representation-and-reasoning}. It enables the precise and structured representation of information, allowing reasoning processes to be modelled algorithmically \cite{ai-mordern}. 

Traditionally, reasoning algorithms were monotonic, which meant that adding new information to a domain did not invalidate previously established conclusions \cite{adam_dis}. However, monotonicity presented challenges when modelling the characteristics of a domain \cite{problem-monotonic} where incoming information may have contradicted or served as an exception to the existing rules.

Nonmonotonic reasoning (NMR) aims to address these limitations by incorporating mechanisms to handle exceptions \cite{nonmontonic_reasoning_definition}. This involves making and revising assumptions previously made when new information is learned. NMR is based on the notion of \textit{typicality}, which means that not all properties of a given object necessarily hold \cite{handbook-history-of-logic}. For example, instead of asserting that ``all birds fly'', we say that ``birds \textit{typically} fly'' as there are exceptional cases such as ostriches and penguins. By defining the characteristics of an object to ``\textit{typically}'' hold, we make provisions for exceptions that could arise. 

Defeasible reasoning offers a framework for modelling this type of reasoning.
Among the various approaches to defeasible reasoning, the KLM framework developed by Kraus, Lehmann, and Magidor \cite{kraus_lehmann_magidor_1990} is well-known for defining the fundamental properties any defeasible reasoning framework should follow when dealing with incomplete information \cite{conditional-knowledge-base-entails}.

A notable reasoning pattern that satisfies the KLM properties is Rational Closure (RC). RC is a conservative form of defeasible reasoning, meaning that it infers minimal assumptions from the knowledge base \cite{casini_meyer_varzinczak_2018}. RC is defined both semantically and algorithmically.

While both KLM and ASP are used to model nonmonotonic reasoning, they make use of different paradigms when it comes to modelling nonmontonic reasoning. The KLM framework adopts a preferential semantics approach, ranking all possible worlds according to a preference relation. A formula $\alpha$ is considered satisfiable if it holds in all of the most preferred models \cite{shoham_1}. ASP employs negation as failure, a form of default reasoning in which a formula $\alpha$ is assumed to be false if it cannot be proven to be true by the given rules within the program. This distinction highlights a key divergence in how each system handles uncertainty and incomplete information. 

Most of the existing approaches for computing entailment relations that satisfy the KLM properties have been imperative \cite{casini_meyer_varzinczak_2018}, with some work exploring declarative programming techniques \cite{declarativeRC}. 

In this paper, we present an alternative declarative definition of RC using ASP. We provide the encoding of RC and prove the correctness of our definitions. Furthermore, we present our findings from testing the computational runtime of our RC definition. Through extensive experiments, we show that our ASP-based implementation outperforms the InfOCF solver \footnote{https://github.com/jonasphilipp/InfOCF}.

\section{Background}
In this section, we provide the relevant background material. Propositional logic is presented in Section \ref{section:propositional-logic}. Defeasible reasoning and the KLM approach are presented in Section \ref{section:defeasibleReasoning}. Rational Closure and Answer Set Programming are presented in Section \ref{section:rational-closure} and \ref{section:answer-set-programming}. 
\subsection{Propositional Logic}
\label{section:propositional-logic}
The finite set of all propositional atoms is denoted by $\mathcal{P}$. Propositional atoms are assigned a \textit{truth value}, either true or false \cite{ben-ari_2012}, and are combined with boolean operators to create propositional statements, otherwise known as formulas. Formulas are denoted by Greek letters such as $\alpha$, $\beta$, and $\gamma$. The language, denoted as $\mathcal{L}$, is a set of formulas which are represented as, for any given $p \in \mathcal{P}$ and $\tau, \alpha \in \mathcal{L}$, $\tau$ can take on the following form: $p, \neg \tau, \tau \wedge \alpha, \tau \rightarrow \alpha, \tau \leftrightarrow \alpha, \tau \vee \alpha$ \cite{ben-ari_2012}. A knowledge base, $\mathcal{K} \subseteq \mathcal{L}$, is a set of propositional formulas which can be finite or infinite. In this paper, we only consider finite knowledge bases. 
We define an interpretation $\mathcal{\omega}$, as a mapping from propositional atoms to truth values. An interpretation, $\mathcal{\omega}$ is defined as $\mathcal{\omega} : \mathcal{P} \rightarrow \{T,F\}$ \cite{ben-ari_2012}. 
We denote the value of a formula under an interpretation as $\mathcal{I}(\alpha)$. If $\mathcal{I}(\alpha)$ is true for some formula $\alpha$, we say that $\mathcal{I}$ satisfies $\alpha$, denoted as $\mathcal{I} \Vdash \alpha$. If at least one interpretation, $\mathcal{I}$, satisfies $\mathcal{K}$ then $\mathcal{I}$ is a \textit{model} of $\mathcal{K}$. The set of all models of $\mathcal{K}$ is denoted as $\text{mod}(\mathcal{K})$. If there exists no $\mathcal{I} \in \mathcal{U}$ such that $\mathcal{I} \Vdash \mathcal{K}$, then $\mathcal{K}$ contains no interpretation satisfying all $\alpha \in \mathcal{K}$. If there is no $\mathcal{I}$ such that $\mathcal{I} \Vdash \mathcal{K}$, $\mathcal{K}$ is said to be unsatisfiable. Drawing inferences from a knowledge base is defined as entailment, denoted as $\models$. 

\subsection{Defeasible Reasoning}
\label{section:defeasibleReasoning}
Defeasible reasoning is a nonmonotonic framework that allows the retraction of previously established rules when new, contradictory information is introduced to a knowledge base \cite{defeasible}.
Various formulations have been developed to model defeasible reasoning, including belief revision \cite{belief}, default logic \cite{defeault-logic}, and the preferential approach \cite{preferential-logics}. The preferential approach has been extended into what is known as the KLM framework \cite{kraus_lehmann_magidor_1990}. This framework is the paper's primary focus due to its well-defined properties and computationally efficient reasoning algorithms, one of which is rational closure (in short, RC) \cite{casini_meyer_varzinczak_2018}.

Our work is situated in the  \textit{KLM framework}, which extends the propositional language by introducing defeasible implications \cite{conditional-knowledge-base-entails}. A defeasible implication is denoted by $\mid \!\sim $, expressed as $\alpha \mid \! \sim \beta$ where $\alpha, \beta \in \mathcal{L}$. We thus define the language $\mathcal{L}_p = \mathcal{L} \cup \{ \alpha\mid\!\sim\beta \ | \alpha,\beta \in \mathcal{L}\}$ to be an extension of the propositional language, $\mathcal{L}$, with the added defeasible connective. This allows us to form formulas using $\mid \!\sim$. Note that we do not allow nesting of $\mid\!\sim$.

Drawing inferences from a set of defeasible implications is called defeasible entailment. The notion of defeasible entailment ($\mid\! \approx$) is not unique; many different ways of defining defeasible entailment have been outlined. The KLM definition of defeasible entailment is of interest in this paper due to its properties, the KLM properties \cite{kraus_lehmann_magidor_1990}, and well-defined algorithms associated with it. 

\subsection{Rational Closure}
\label{section:rational-closure}
RC is a conservative form of defeasible reasoning, which means that the reasoning algorithm infers as minimally as possible from a knowledge base \cite{adam_dis} but allows for handling exceptional information. The framework is defined semantically as well as algorithmically \cite{casini_meyer_varzinczak_2018}. This paper will focus on the algorithmic definition of RC as outlined by Casini et al. \cite{casini_meyer_varzinczak_2018}. Computing defeasible entailment with RC requires two algorithms, BaseRank and RC. 

\begin{algorithm}
\caption{BaseRank}\label{alg:BaseRank}
\label{tab:BaseRank}
\begin{algorithmic}[1]
\State \textbf{Input:} A knowledge base $\mathcal{K}$
\State \textbf{Output:} An ordered tuple ($R_0$,..., $R_{n-1}$,$R_{\infty}$,$n$). Each $R_i$, where $i \in  [0, n]$ indicates the rank each statement is placed on. 
\State $i=0$;
\State $E_0 = \overrightarrow{\mathcal{K}}$; 
\While{$E_{i-1} \neq E_i$} 
    \State $E_{i+1}= \{\ \alpha \rightarrow \beta \in E_i| E_i \models \neg \alpha \}\ $;
    \State $R_{i} = E_i \setminus E_{i+1}$;
    \State $i = i+1$;
\EndWhile
\State $R_\infty = E_{i-1}$;
\If {$E_{i-1} = \emptyset$}
    \State $n = i-1$;
\Else 
\State $ n= i$;
\EndIf
\State \textbf{return} ($R_0$,..., $R_{n-1}$,$R_{\infty}$,$n$);
\end{algorithmic}
\end{algorithm}

BaseRank is a procedure which assigns a numerical rank to statements in a knowledge base based on their exceptionality. General statements, which apply broadly to a particular topic, are given lower ranks, while more exceptional statements which violate established rules within a knowledge base are assigned higher ranks. For example, given the knowledge base $\mathcal{K} = \{ \mathrm{man} \mid\!\sim \mathrm{mortal},\ \mathrm{Socrates} \rightarrow \mathrm{man}, \ \mathrm{Socrates} \rightarrow \neg \mathrm{mortal} \} $, the statement $\mathrm{Socrates} \rightarrow \neg \mathrm{mortal}$ is exceptional while $\mathrm{man} \entails \mathrm{mortal}$ is a general statement applicable to the class \emph{man}. 

The BaseRank procedure performs ranking on a knowledge base which has been transformed to only classical material implications. We refer to this as materialisation, formalised in Definition \ref{material}. Algorithm \ref{tab:BaseRank} outlines the BaseRank procedure. 

\begin{ax}
\label{material}
\textnormal{The materialisation of a knowledge base $\mathcal{K}$ is defined as $\overrightarrow{\mathcal{K}}$:= $\{ \alpha \rightarrow \beta \ | \ \alpha$  $\mid \! \sim \beta \in \mathcal{K}\}$ \cite{adam_dis}.}
\end{ax}
\begin{ax} 
\label{exceptional}
\textnormal{A formula $\alpha$ is exceptional in the set $\mathcal{K}$ if $\mathcal{K}\models\neg \alpha$ \cite{adam_dis}.}
\end{ax}

RC determines whether a knowledge base entails an implication. For a given query $\alpha \entails  \beta$, $\alpha$ is the antecedent and $\beta$ is the consequence. To compute defeasible entailment for a given query, $\alpha \entails \beta$, RC takes a defeasible knowledge base, computes BaseRank, then repeatedly eliminates ranks assigned to statements within the knowledge base which make the antecedent exceptional, eventually settling on the maximal subset $\overrightarrow{\mathcal{K'}}$ of $\overrightarrow{\mathcal{K}}$ which does not make $\alpha$ exceptional. RC returns true to a query if $\alpha \rightarrow \beta$ is classically entailed by $\overrightarrow{\mathcal{K'}}$. Entailment under RC is denoted as $\mathcal{K} \mid \! \approx_{RC}$. Algorithm \ref{alg:RationalClosure} outlines the RC procedure. 

\begin{algorithm}
\caption{Rational Closure}\label{alg:RationalClosure}
\begin{algorithmic}[1]
\State \textbf{Input:} A knowledge base $\mathcal{K}$ and a defeasible implication $\alpha \entails \beta$.
\State \textbf{Output:} \textbf{true}, if the query is entailed by the knowledge base, else \textbf{false}.
\State ($R_0$,..., $R_{n-1}$,$R_{\infty}$,$n$) = $BaseRank(\mathcal{K})$;
\State $i=0$; 
\State $\mathcal{R} = \bigcup^{j<n-1}_{j=0} \mathcal{R}_j$;
\While{$\mathcal{R}_\infty \cup \mathcal{R} \models \neg \alpha \ and \ \mathcal{R} \neq \emptyset$} 
    \State $\mathcal{R} = \mathcal{R} \setminus \mathcal{R}_i$;
    \State $i = i+1$;
\EndWhile
\State \textbf{return} $\mathcal{R}_\infty \bigcup \mathcal{R} \models \alpha \rightarrow \beta$;
\end{algorithmic}
\end{algorithm}
\begin{example}\label{ex:Rational Closure Example}
  Given the knowledge base $\mathcal{K}_{1} = \{ \mathrm{man} \mid\!\sim \mathrm{mortal},\ \mathrm{Socrates} \rightarrow \mathrm{man}, \ \mathrm{Socrates} \rightarrow \neg \mathrm{mortal} \} $ and the query, $\mathrm{Socrates} \rightarrow \mathrm{man}$. The antecedent of the query is $\mathrm{Socrates}$. The materialisation of $\mathcal{K}_{1}$ is $\{ \mathrm{man} \rightarrow \mathrm{mortal},\ \mathrm{Socrates} \rightarrow \mathrm{man}, \ \mathrm{Socrates} \rightarrow \neg \mathrm{mortal} \} $ and the base rank for $\mathcal{K}_{1}$ is outlined in Table \ref{tab:two_column_table}. 

  \begin{table}[h!]
    \centering
    \begin{minipage}{0.45\textwidth}
 
        \centering
        \captionof{table}{BaseRank of knowledge base $\mathcal{K}_{1}$} 
 \label{tab:two_column_table}
        \vspace{0.3em}
        \begin{tabular}{|c|c|}
            \hline
            $\mathcal{R_\infty}$ & $\mathrm{Socrates} \rightarrow \mathrm{man}, \ \mathrm{Socrates} \rightarrow \neg \mathrm{mortal}$ \\
            \hline
            $\mathcal{R}_0$ & $\mathrm{man} \rightarrow \mathrm{mortal}$ \\
            \hline
        \end{tabular}
    \end{minipage}
    \hfill
    \begin{minipage}{0.45\textwidth}
    
        \centering
 \captionof{table}{BaseRank of knowledge base $\mathcal{K}_{1}$ with the lowest rank eliminated.}
\label{tab:two_column_table_lined}
        \vspace{0.3em}
        \begin{tabular}{|c|c|}
            \hline
            $\mathcal{R_\infty}$ & $\mathrm{Socrates} \rightarrow \mathrm{man}, \ \mathrm{Socrates} \rightarrow \neg \mathrm{mortal}$ \\
            \hline
            $\mathcal{R}_0$ & \hcancel{$\mathrm{man} \rightarrow \mathrm{mortal}$} \\
            \hline
        \end{tabular}
    \end{minipage}
\end{table}

 To determine if $\mathcal{R} \cup \mathcal{R}_{\infty} \models \neg \mathrm{Socrates}$;  models of $\mathcal{K}_{1}$ are computed. We denote Socrates as $s$, man as $m$, and mortal as $w$. The models of $\mathcal{K}_{1}$ are \textit{mod}$(\mathcal{K}_{1}) = \{\overline{s}mw,\overline{sm}w,\overline{smw}\}$. We note that for all interpretations of $\mathcal{K}_{1}$ $\neg \mathrm{Socrates}$ holds, $\mathcal{R} \cup \mathcal{R}_{\infty} \models \neg \mathrm{Socrates}$. RC will thus eliminate rank 0. The revised models of $\mathcal{K}_{1}$ are  $\{\overline{s}mw,\overline{sm}w,\overline{smw},sm\overline{w}\}$. We see that $\mathcal{R} \cup \mathcal{R}_{\infty} \not \models \neg \mathrm{Socrates} $ as there exists an interpretation which makes $\mathrm{Socrates}$ true. Computing classical entailment on $\mathcal{R} \cup \mathcal{R}_{\infty}$, we see that at least one model within the remaining ranks makes the statement $Socrates \rightarrow man$ true, thus under RC, $\mathcal{K} \mid \! \approx_{RC} \mathrm{Socrates} \rightarrow man$.
This example highlights the use of RC. Under classical logic, it would have been concluded that Socrates does not exist. RC allowed us to capture the exceptional case of Socrates and represent this within a knowledge base. 
\end{example}

\subsection{Answer Set Programming}
\label{section:answer-set-programming}
ASP is a modern approach to declarative programming designed to solve complex problems \cite{answer_set_programming}. It is recognised for its ease of use, expressive constructs, and computational efficiency \cite{answer_set_practice}.
ASP consists of rules similar to those in Prolog \cite{prolog} and has computational methods that leverage the concepts behind efficient satisfiability solvers \cite{answer_set_practice}. Solving ASP programs involves determining stable models \cite{12_defintions_of_stable_models}. A stable model is a set of atoms which satisfy the rules specified in an ASP program. Problems are encoded into programs that can be processed by ASP tools like \textit{gringo} \cite{gringo} and \textit{clasp} \cite{clasp}. The solver used in this paper is \textit{clasp} \cite{clasp}. \textit{Clingo} \cite{answer_set_practice} combines both \textit{gringo} and \textit{clasp} into a monolithic system and is a programming tool based on the ASP programming paradigm.
Rules in ASP take the form: 
$
a_{0} \leftarrow a_{1}, a_{2},..., a_{m}, \sim a_{m+1},.., \sim a_{n}. 
$
We denote its head as \( \text{head}(r) = \{ a_0 \} \). 
The body is defined as \( \text{body}(r) = \{a_{1}, a_{2}, \dots ,a_{m}, \sim a_{m+1}, \dots , \sim a_{n}\} \).
 Both the head and the body of the rule consist of literals. Literals are atoms that are either positive or negative \cite{answer_set_practice}. The set of positive literals in the body is denoted by $body^{+}(r) = \{ a_{1}, a_{2},..., a_{m} \} $ while the set of negative literals is $body^{-}(r) = \{ a_{m+1},..., a_{n} \} $. Intuitively, if we have a set $\mathcal{X}$ which consists of literals within the program then if $body^{+}(r) \subseteq \mathcal{X}$ and $body^{-}(r) \cap \mathcal{X} = \emptyset$ then $head(r) \in \mathcal{X}$ \cite{answer_set_practice}. Answer sets are minimal supported models of an ASP program. The full definition can be found in \cite{answer_set_practice}.

\emph{Choice rules} describe several alternative ways to form a stable model, with the head of the rule consisting of an expression in braces \cite{answer_set_practice}. For example:
  $  \{a;b\}$

states all possible ways the two literals $a$ and $b$ can be included in the stable model. There are four possible choices, which are the power set of $\{ a,b \}$, namely,
 $   \{ \{\emptyset\}, \{a\}, \{b \}, \{a,b\}\}.$

ASP allows for expressing cost functions subject
to minimisation and/or maximisation. This is especially useful when solving optimisation problems. The function we are interested in is the \textit{minimize statement}, expressed as \textit{minimize\{$l_1 = w_1@p_1, \dots , l_n = w_n@p_n$\}}. The minimize directive instructs the solver to compute the optimal answer set that minimises the elements' weighted sum, $ l_1 = w_1@p_1, \dots , l_n = w_n@p_n $. 

\section{Declarative approach to BaseRank and Rational Closure} 
\label{section:asp-def-rc-and-base-rank}
In this section, we present our definition of BaseRank (Section \ref{section:base-rank}) and RC (Section \ref{section:RationalClosureASP}). In each, we describe the structure of the problem instance and its corresponding problem encoding in ASP.

\subsection{BaseRank}
\label{section:base-rank}

We describe classical implications ($\alpha \rightarrow \beta$) and defeasible implications ($\alpha \mid \! \sim \beta$) by the following predicates, $\mathtt{classical}/2$ and $\mathtt{defeasible}/2$. These predicates serve as the problem instances for BaseRank. 
Listing \ref{fig:problem-instance-base-rank} shows how we would view the knowledge base $\mathcal{K}_1$ as presented in Example \ref{ex:Rational Closure Example}. 
\begin{lstlisting}[style=asp,caption={Encoding of knowledge base $\mathcal{K}_1$}, label=fig:problem-instance-base-rank]
defeasible(man,mortal).
classical(socrates,man).
classical(socrates,-mortal).
\end{lstlisting}

The problem encodings for BaseRank are solved via schematic rules. Schematic rules contain variables and allow for the generalisation of logic rules \cite{answer_set_practice}. Our definition of BaseRank in Listing \ref{fig:problem-encoding-base-rank} follows the \textit{generate, define and test} methodology \cite{generateMethodology}. 

The preliminary rules on lines 2 and 3 convert the knowledge base encodings to material implications, defined by the predicate $\mathtt{m\_implication/2}$. All classical rules are then ranked on the infinite rank, as described by Line 4. 

To ensure the knowledge base itself is not inconsistent, we define the predicate $\mathtt{infer}/2$ on lines 5,6 and 7. Any knowledge base which derives $\top \rightarrow \alpha$ and $\neg \alpha \rightarrow \alpha$ would be unsatisfiable; thus, the $\mathtt{infer/2}$ predicate allows us to define constraints to catch these cases and define a global state of the knowledge base.  

The \textit{generate} section on line 8 arbitrarily assigns a numerical rank to every material implication using a choice rule. The predicate $\mathtt{rank(m\_implication(X,Y),N)}$ expresses that the material implication, $X \rightarrow Y $ encoded as $\mathtt{m\_implication(X,Y)}$ is assigned a numerical rank $\mathtt{N}$. The choice rule has a cardinality constraint of one. This ensures that for every solution, a material implication is placed on one rank. 

In the worst case, each statement will be placed on a unique rank. Thus, the choice rule on line 8 assigns each material implication to an arbitrary rank, where the maximum rank a statement can be placed on is based on the number of statements within a knowledge base. For instance, if we consider $\mathcal{K}_1$, in the worst case, each statement would be assigned a unique rank, meaning there are 3 ranks for the ranked model.

The \textit{define} section of the encoding is from lines 9 to 12. This section derives all the possible consequences within the knowledge base. The $\mathtt{derive/3}$ predicate encodes the natural deductive rules for transitivity, identity and disjunctive elimination. The predicate $\mathtt{derive(X,Y,N)}$ is defined as $X\rightarrow Y$ holds on rank $\mathtt{N}$. 

The \textit{test} section of the encoding is from lines 14 to 19.  
 Line 14 ensures that if an answer set contains the predicate $\mathtt{derive(X,Y,N)}$ and the $\mathtt{derive(X,-Y,N)}$, the answer set is rejected. Line 15 ensures that any antecedent which is a tautology should not have the negation of itself present on that rank. Lines 18 and 19 ensure that inconsistent knowledge bases are rejected. 

Lines 21 to 23 make use of the $\mathtt{\#minimize}$ optimisation directive to ensure all implications are assigned the lowest ranks possible while preserving the distribution structure of implications; furthermore, all antecedents which are tautologies are assigned the lowest ranks possible.

\begin{lstlisting}[style=asp,caption={Problem encoding of BaseRank}, label=fig:problem-encoding-base-rank]
m_implication(X,Y) :- defeasible(X,Y).
m_implication(X,Y) :- classical(X,Y).
statement_count(C):- C = #count{X,Y : m_implication(X,Y)}.
rank(m_implication(X,Y),inf) :- classical(X,Y).
infer(X,Y) :- rank(m_implication(X,Y),N).
infer(X,P) :- rank(m_implication(Y,P),N1), infer(X,Y).
infer(top,Y) :- infer(X,Y), infer(-X,Y).
{rank(m_implication(X,Y),0..C-1)} =1 :- statement_count(C), m_implication(X,Y), not classical(X,Y).
derive(X,X,N) :- rank(m_implication(X,Y),N).
derive(X,Y,N) :- rank(m_implication(X,Y),N), derive(X,X,N).
derive(X,P,N) :- rank(m_implication(Y,P),N1), derive(X,Y,N), N1 >= N.
derive(top,Y,N) :- derive(X,Y,N1), derive(-X,Y,N) , N1 >= N.

:- derive(X,Y,N), derive(X,-Y,N).
:- derive(-X,_,N), derive(top,X,N).
:- derive(_,-X,N), derive(top,X,N).
:- rank(m_implication(X,_),N1), rank(m_implication(X,_),N2), N1 != N2, not classical(X,_), not coded_classical(X,_).
:- infer(top,X), infer(-X,X). 
:- infer(top,Z), infer(top,X), infer(Z,-X). 

#minimize{N,X,Y : rank(m_implication(X,Y),N)}.
#minimize{N,Y : derive(top,Y,N)}.
#minimize{N,Y,X : derive(top,Y,N),rank(m_implication(-X,Y),N) , rank(m_implication(X,Y),N)}.
#show rank/2.


\end{lstlisting}
The computed ranked model for $\mathcal{K}_1$ using the BaseRank definition as defined in Listing $\ref{fig:problem-encoding-base-rank}$ is shown in Listing \ref{fig:problem-rank}.

\subsection{Rational Closure}
\label{section:RationalClosureASP}
A query is defined via the predicate $\mathtt{query(X,Y)}$, while the ranked levels of the statements obtained from computing BaseRank are described via the predicate $\mathtt{rank(m\_implication(X,Y),N)}$. Our definition of RC can handle queries of the form $p \entails q$ and $p$ where $p, q$ are literals and contain no boolean connectives. 

\begin{lstlisting}[style=asp,caption={BaseRank model of $\mathcal{K}_1$}, label=fig:problem-rank]
rank(m_implication(socrates,man),inf).
rank(m_implication(socrates,-mortal),inf).
rank(m_implication(man,mortal),0).
\end{lstlisting}
Similar to BaseRank, the definition of RC follows the \textit{ generate, define and test} approach.

In Line 1 of Listing \ref{fig:problem-encoding-rational-closure}, we generate the predicate $\mathtt{guess}/1$ by means of a choice rule. This rule states that for all the rank levels $\mathtt{N}$ that exist in the $\mathtt{rank(_, N)}$ facts, choose exactly one guess(N). The \textbf{$\mathtt{guess}/1$} predicate acts as an arbitrary rank chosen from the ranked model. 

The predicate $\mathtt{inference}/2$ encodes material implications that can be inferred by the knowledge base. These are outlined by the rules in Lines 4-8. Line 4 assumes some interpretation exists which makes the antecedent $\mathtt{X}$ true; this follows to be the identity rule of natural deduction; thus, the fact $\mathtt{inference(X, X)}$ is added to the answer set. Line 6 makes use of the transitive rule for material implications, and Lines 8 and 9 capture all logical inferences from the material implications by making use of contrapositive rules.

\begin{lstlisting}[style=asp, caption={Problem encoding of Rational Closure}, label={fig:problem-encoding-rational-closure}]
{guess(N):rank(_,N)}=1.
inference(X,X) :- rank(m_implication(X,Y),N1), guess(N), N<=N1.
inference(X,Y) :- rank(m_implication(X,Y),N1), guess(N), N<=N1.
inference(X,Z) :- inference(X,Y), rank(m_implication(Y,Z),N1), guess(N), N<=N1.
inference(-Y,-X) :- rank(m_implication(X,Y),N1), guess(N), N<=N1.
inference(-Z,-X) :- inference(X,Y), rank(m_implication(Y,Z),N1), guess(N), N<=N1.
inference(top,Y) :-inference(X,Y), inference(-X,Y).
inference(top,-X) :-inference(X,Y), inference(X,-Y).
:- query(X,Y), inference(top, -X).
entailed(true):- inference(X,Y), query(X,Y).
entailed(false) :- not entailed(true).
#minimize{N:guess(N)}.
#show entailed/1.
\end{lstlisting}

Lines 7 and 8 play vital roles in determining which ranks satisfy the query antecedent. Both lines encode disjunctive elimination. For line 7, if we have some inference, for example,  $\mathtt{inference(X,Y)}$ and $\mathtt{inference(-X,Y)}$ then it is always true that the consequence, $\mathtt{Y}$, holds. Line 8 states that if we can derive $\mathtt{inference(X,Y)}$ and $\mathtt{inference(X,-Y)}$ then $-X$ is true, making $X$ false. 

The inference rules correspond to standard natural deduction rules, namely, identity, modus ponens, contrapositive, and disjunctive elimination.

Line 9 rejects any answer set where the antecedent of a query is exceptional in a guessed rank. Entailment is defined on line 10. This definition states that if $\mathtt{inference(X,Y)}$ is defined in a given answer set and $\mathtt{query(X,Y)}$ also holds, then the query is entailed by the knowledge base. A statement is not entailed by a knowledge base when there is no evidence in the answer set to suggest that $\mathtt{entailed(true)}$ holds. 

The $\mathtt{\#minimize}$ directive ensures the smallest rank, which does not result in an exceptional antecedent, is chosen. 

\begin{example}
    We begin by defining a simple knowledge base, $\mathcal{K}_1$, shown in Listing~\ref{fig:problem-instance}. This includes defeasible and classical conditionals as well as a query. The knowledge base is encoded in a file called \texttt{knowledge\_base.lp}.

\begin{lstlisting}[style=asp,caption={Encoding of knowledge base $\mathcal{K}_1$}, label=fig:problem-instance] 
defeasible(man,mortal).
classical(socrates,man). classical(socrates,-mortal).
query(socrates,man).
\end{lstlisting}

To evaluate the query using the RC procedure, we run the following command in our terminal \texttt{clingo knowledge\_base.lp base\_rank.lp rational\_closure.lp}.
The resulting optimal answer set is shown in Listing~\ref{fig:aspRC}. The output given shows that only one answer set was computed by the solver. We observe that the rank guess was the infinite rank, which led to the inferences shown in the answer sets. Given that the atom $\mathtt{inference(socrates,man)}$ is included in the answer set using the inference rules, the query $\mathtt{query(socrates,man)}$ is entailed by the knowledge base.

\begin{lstlisting}[style=asp,caption={RC entailment results using Clingo for $\mathcal{K}_1$}, label=fig:aspRC] 
query(socrates,man) guess(inf) 
inference(socrates,socrates) inference(-man,-socrates) 
inference(mortal,-socrates) inference(socrates,man)
inference(socrates,-mortal) entailed(true)
\end{lstlisting}
\end{example}

\subsection{Soundness and Correctness}
To prove the correctness of our ASP definitions of BaseRank and RC, we prove the soundness and completeness of each. 
\subsubsection{BaseRank}
First, some notational preliminaries. We denote, given a knowledge base ${\cal K}$, $\mathtt{encoding}({\cal K})$ be its encoding (see Section \ref{tab:BaseRank}) and $\mathtt{baseRank}$ the program in Listing \ref{fig:problem-encoding-base-rank} (excluding the minimize-directive on lines 22-24). We denote the minimize-directive on lines 22-24 of Listing \ref{fig:problem-encoding-base-rank} by $\mathtt{baseRankMin}$.
Furthermore, for a given answer set ${\cal S}$ of $\mathtt{encoding}({\cal K})\cup \mathtt{baseRank}$, we define ${\cal K}^{\cal S}_i$ and $\overrightarrow{{\cal K}}^{\cal S}_i$ as follows: \[
{\cal K}^{\cal S}_i= \{ A\entails B\in {\cal K}\mid \mathtt{rank(m\_implication(A,B),i)}\in {\cal S}\}
\] \[\
\overrightarrow{{\cal K}}^{\cal S}_i = \{ A\rightarrow B\mid A\entails B\in {\cal K}^{\cal S}_i\}\]

The following shows that the derived rules encoded in Base rank are exactly those rules used in natural deduction \cite{predicateLogic}. 
\begin{Propositionrep}
\label{naturalDeductionToASPDefinition}
Let \(i\) be a fixed nonnegative integer and $\cal S$ be some answer set satisfying the rules outlined in $\mathtt{BaseRank}$ without the minimize directive.  
Further define
\[
\mathrm{ND}_N
\;=\;
\bigl\{\,X \to Y 
\;\bigm|\; \overrightarrow{{\cal K}}^{\cal S}_i \models X\rightarrow Y
\bigr \}\]
and
\[
\mathrm{ASP}_i
\;=\;
\bigl\{\,X \to Y 
\;\bigm|\;
\mathtt{derive(X,Y,i) \in \cal S}\bigr\}.
\]
Then
\[
\mathrm{ND}_i \;=\; \mathrm{ASP}_i.
\]

\end{Propositionrep}

\begin{appendixproof}
First note that, as our language is nothing but “if–then” between literals, the only non-trivial reasoning steps are identity, modus ponens, and disjunctive elimination.

We proceed by showing the following: 
\[
\mathrm{ND}_N \;\subseteq\; \mathrm{ASP}_N
\quad\text{and}\quad
\mathrm{ASP}_N \;\subseteq\; \mathrm{ND}_N.
\]

\medskip

\noindent\textbf{(1) \(\mathrm{ND}_N \subseteq \mathrm{ASP}_N\)}
 Let 
\[
X \;\to\; Y 
\;\in\; \mathrm{ND}_N.
\]
By definition, there is a natural deduction proof of \(X \to Y\) from premises in \(\mathcal{K}_N\), 
We refer to this proof as \(\Pi\), which has length n.  We show that \(\texttt{derive}(X,Y,N)\) $\in$ ASP\_N.  We proceed by structural induction on the length of deductive proof  \(\Pi\).

\medskip

\noindent\emph{Base Case (n=0)}: 
If n=0, then this is simply the identity rule and $X \rightarrow Y$ is a premise in $\mathcal{K}_N$, from rule 
\[
\texttt{derive(X,Y,N)} \;:-\; \texttt{rank(m\_implication(X,Y),N) , derive(X,X,N).}
\]
is added into the answer set, this $\mathtt{derive(X,Y,N)} \in \cal S$, $X\rightarrow Y  \in ASP_{N}$. 
\newline
\noindent\emph{Inductive Hypothesis (n=k):} 
For any natural deductive proof \(\Pi\) of length k if $X\rightarrow Y \in ND_N$ then $X \rightarrow Y \in ASP_N$ 
\newline

\noindent\emph{Inductive Step (\(n = k+1\)).}  
Let \(\Pi\) be a natural‐deduction proof of \(X \to Y\) of length \(k+1\).  We consider the following cases:

\begin{description}
  \item[\({Idenity}:\)]  
    In this case, the last step is “reiterate the premise \(X \to Y\).”  In natural deduction notation:
    \[
      \frac{ }{\,X \to Y \;\vdash\; X \to Y\,}
      \quad\Longrightarrow\quad
      X \to Y.
    \]
    Since \(X \to Y \in \mathcal{K}_N\), the corresponding ASP clause
    \[
      \mathtt{derive}(X,Y,N) 
        \;:-\; \mathtt{rank}\bigl(\mathtt{m\_implication}(X,Y),N\bigr)
    \]
    fires, so \(\mathtt{derive}(X,Y,N)\in \mathcal{S}\).  Hence \(X \to Y \in \mathrm{ASP}_N\).

  \item[\({Modus \  Ponens \  / \  Transitivity}:\)]  
    \[
      \frac{\,X \to Z \quad Z \to Y\,}{\,X \to Y\,}
    \]
    By inductive hypothesis, both \(X \to Z\) and \(Z \to Y\) yield
    \(\mathtt{derive}(X,Z,N)\in \mathcal{S}\) and \(\mathtt{derive}(Z,Y,N)\in \mathcal{S}\).  
    Then the ASP rule
    \[
      \mathtt{derive}(X,Y,N) 
        \;:-\; \mathtt{derive}(X,Z,N),\; \mathtt{derive}(Z,Y,N)
    \]
    applies, giving \(\mathtt{derive}(X,Y,N)\).  Thus \(X \to Y \in \mathrm{ASP}_N\).

  \item[\({Disjunction \ Elimination}:\)]  
    \[
      \frac{
        [\,X\,]\;\vdash\; Y
        \quad
        [\,\neg X\,]\;\vdash\; Y
      }{\,Y\,}
    \]
    By the inductive hypothesis, we have
    \(\mathtt{derive}(X,Y,N)\in \mathcal{S}\) and \(\mathtt{derive}(-X,Y,N)\in \mathcal{S}\).  
    Then ASP’s disjunction‐elimination clause
    \[
      \mathtt{derive}(\top, Y, N) 
        \;:-\; \mathtt{derive}(X,Y,N),\; \mathtt{derive}(-X,Y,N)
    \]
    adding \(\mathtt{derive}(\top,Y,N)\) into the answer set, i.e.\ \(Y \in \mathrm{ASP}_N\).
\end{description}

In every case, \(\mathtt{derive}(X,Y,N)\) (or \(\mathtt{derive}(\top,Y,N)\)) appears in \(\mathcal{S}\).  Hence \(\mathrm{ND}_N \subseteq \mathrm{ASP}_N\).

\medskip
\noindent\textbf{(2) \(\mathrm{ASP}_N \subseteq \mathrm{ND}_N\).}

Conversely, suppose \(\mathtt{derive}(X,Y,N)\in \mathcal{S}\). By inspecting the BaseRank encoding, there are exactly three ways \(\mathtt{derive}(X,Y,N)\) can be derived:
\begin{enumerate}
  \item $\mathtt{derive(X,Y,N)}$ is directly derived from $\mathtt{rank(m\_implications(X,Y),N)}$, line 10 from Listing \ref{fig:problem-encoding-base-rank}. 
  \item $\mathtt{derive(X,Y,N)}$ is derived from \(\mathtt{derive(X,Z,N)}\) and \(\mathtt{derive(Z,Y,N)}\), line 11 from Listing \ref{fig:problem-encoding-base-rank}.
  \item $\mathtt{derive(top,Z,N)}$ is derived from \(\mathtt{derive(X,Z,N)}\) and \(\mathtt{derive(-X,Z,N)}\), line 12 from Listing \ref{fig:problem-encoding-base-rank}.
\end{enumerate}
We reconstruct a natural deduction proof of \(X\to Y\) by induction on the ASP‐derivation depth of the derive predicate, \(\mathtt{derive(X,Y,N)}\). 
\newline
\noindent\emph{Base Case (n=0)}: 
If the length of the derive predicate is 0 then  \(\mathtt{derive}(X,Y,N)\) is added into the answer set $\cal S$ from 
    \(\mathtt{rank}(\mathtt{m\_implication}(X,Y),N)\), thus\(X \to Y \in \mathcal{K}_N\) and $X \rightarrow Y \in ND_N$.

\noindent\emph{Inductive hypothesis (n=k)}: 
For length n=k, then for every $\mathtt{derive(X,Y,N) \in \cal S}$, for every $X \to Y \in ASP_N$, $X \to Y \in ND_N$. 

\noindent\emph{Inductive Step (n=k+1)}: 
\begin{itemize}

  \item If \(\mathtt{derive(X,Y,N)}\) was obtained from $\mathtt{rank(m\_implications(X,Y),N)}$ then this is simply the base case so this $ X \to Y \in ND_N$. 
  \item If \(\mathtt{derive}(X,Y,N)\) results from 
\(\mathtt{derive}(X,Z,N)\) together with 
\(\mathtt{rank}(\mathtt{m\_implication}(Z,Y),N)\). From the inductive hypothesis, $X\to Z$ and $Z \to Y \in ND_N$, thus through the use of natural deduction, we can show that $ X \to Y \in ND_N$. 
  \item If \(\mathtt{derive}(X,Y,N)\) and \(\mathtt{derive}(-X,Y,N)\) hold then \(\mathtt{derive}(\top,Y,N)\ \in {\cal S} \), by the inductive hypothesis, $X \to Y$ and $X\to \neg Y \in ND_N$, through natural deduction $\top \to Y \in ND_N$. 
\end{itemize}

In each scenario, we build a valid ND subproof for \(X\to Y\).  Hence \(\mathrm{ASP}_N \subseteq \mathrm{ND}_N\).

\medskip

Combining (1) and (2) completes the proof that \(\mathrm{ND}_N = \mathrm{ASP}_N\).
\end{appendixproof}

The following Proposition shows that the rankings assigned by $\mathtt{baseRank}$ respect the notion of tolerance or exceptionality used in the BaseRank-algorithm:

\begin{Propositionrep}
Let a knowledge base ${\cal K}$ be given. For a given answer set ${\cal S}$ of $\mathtt{encoding}({\cal K})\cup \mathtt{baseRank}$, for any $i\geq 0$, and
for any $A\entails B \in{\cal K}^{\cal S}_i$, $\bigcup_{j=i}^\infty  \overrightarrow{{\cal K}^{\cal S}_j}\not\models \lnot A$.
\end{Propositionrep}
\begin{appendixproof}
    Assume by contradiction that for some $A\entails B \in{\cal K}^{\cal S}_i$, $ \bigcup_{j=i}^\infty  \overrightarrow{{\cal K}^{\cal S}_j}\models \lnot A$. This entails that every interpretation that satisfies $\bigcup_{j=i}^{\infty}\overrightarrow{\mathcal{K}_j^{\mathcal{S}}}$ also satisfies $\neg A$. From the equivalence between the ASP rules and natural deductive rules from \ref{naturalDeductionToASPDefinition} we see that this means $\mathtt{derive(top,-A,j)}$ is added into the answer set \cal S. From this we see that using natural deduction rules, $\bigcup_{j=i}^{\infty}\overrightarrow{\mathcal{K}_j^{\mathcal{S}}} \models (A \rightarrow \bot).$ holds. However, given that  $A\entails B \in{\cal K}^{\cal S}_i$, this means $\mathtt{derive(A,B,j)} \in \cal S$. From this we see that $\mathtt{derive(A,B,j)}, \mathtt{derive(top,-A,j)}\in \cal S$. This does not hold due to the integrity constraint on line 15 of Listing \ref{fig:problem-encoding-base-rank}. Thus $ \bigcup_{j=i}^\infty  \overrightarrow{{\cal K}^{\cal S}_j}\not\models \lnot A$
\end{appendixproof}

Furthermore, $\mathtt{baseRank}$ is sound in the sense that every ``refinement'' of the base ranks of ${\cal K}$ will occur as an
answer set of $\mathtt{encoding}({\cal K})\cup \mathtt{baseRank}$:
\begin{Propositionrep}\label{prop:baserankeq}
Let a knowledge base ${\cal K}$  with $\mathtt{BaseRank}({\cal K})=({\cal R}_0,\ldots,{\cal R}_n,{\cal R}_\infty)$ be given.
For any partition $({\cal R}'_0,\ldots,{\cal R}'_m,{\cal R}_\infty)$ of  ${\cal K}$ s.t.\ for every ${\cal R}'_i$ there is a $j$ s.t.\ ${\cal R}'_i\subseteq {\cal R}_j$,  there is an answer set ${\cal S}$ of $\mathtt{encoding}({\cal K})\cup \mathtt{baseRank}$ s.t.\ ${\cal K}^{\cal S}_i= {\cal R}'_i$ for every $i\geq 0$.
\end{Propositionrep}
\begin{appendixproof}
Given a refinement $(R'_0, \dots, R'_m, R_\infty)$ of the BaseRank partition $(R_0, \dots, R_n, R_\infty)$, where for each $R'_i$ there exists some $j$ such that $R'_i \subseteq R_j$. Given $R'_i$, we can construct a answer set $\cal S$ as follows:
\begin{itemize}
    \item For each defeasible implication $A \mid\!\sim B \in R'_i$, include the atom \texttt{rank(m\_implication(A,B), i)} in $S$.
    \item For each classical implication $A \rightarrow B \in K$, include the atom \texttt{rank(m\_implication(A,B), inf)} in $S$ (as per the encoding).
    \item Include all atoms derivable from these rankings under the inference and propagation rules of the encoding, $\mathtt{derive/3}$, $\mathtt{infer/2}$.
\end{itemize}

Given this answer set $\cal S$ we will show that it is satisfied by the integrity constraints defined by $\mathtt{BaseRank}$.  

\begin{itemize}
\item \textbf{Integrity constraint on line 14}:\\
This constraint forbids deriving $X \to Y$ and $X \to -Y$ simultaneously at rank $N$. The original BaseRank partition $(R_0,\dots,R_n,R_{\infty})$ is consistent by construction. Since every $R'_i$ is a subset of some $R_j$, the refinement cannot introduce contradictions. Thus, line 18 holds.

\item \textbf{Integrity constraint on lines 15-16}:\\
These constraints forbid deriving both $-X$ when $X$ is a tautology on the level $N$. Again, since the original partition is consistent, any subset is consistent. Thus, the integrity constraint holds. 

\item \textbf{Integrity constraint on lines 17}:\\
The construction explicitly assigns exactly one rank to each implication within the knowledge base, $\mathcal{K}$, thus the integrity constraint holds.
\end{itemize}
Thus, ${\cal K}^{S}_{i} = {\cal R'}_{i}$ for each $i$.
\end{appendixproof}

Finally, we show that the minimal answer set of the $\mathtt{baseRank}$-encoding coincides with the partition generated by the BaseRank algorithm:
\begin{Propositionrep}
\label{baserank:sound}
Let a knowledge base ${\cal K}$ with $\mathit{BaseRank}({\cal K})=({\cal R}_0,\ldots,{\cal R}_n,{\cal R}_\infty)$ be given. 
For the answer set ${\cal S}$ of $\mathtt{encoding}({\cal K})\cup \mathtt{baseRank}\cup\mathtt{baseRankMin}$, ${\cal K}^{S}_{i} = {\cal R}_{i}$ for $i\geq 0$.
\end{Propositionrep}

\begin{appendixproof}
Suppose some defeasible conditional $\delta = \alpha \mid\!\sim \beta$ is placed at rank $j > i$ in the minimal answer set $S$, whereas $\delta \in R_i$ in $\mathit{BaseRank}(K)$. Since $\mathit{BaseRank}$ assigns $\delta$ to the lowest consistent rank, we know that placing $\delta$ at rank $i$ would still yield a consistent interpretation.

Thus, we can construct an alternative answer set $S'$ where $\delta$ is reassigned to rank $i$, keeping all other assignments unchanged. This new answer set would have a strictly lower cost under the $\#\mathit{minimize}$ directive, contradicting the assumption that $S$ is minimal.

 Conversely, suppose there exists a defeasible conditional $\delta \in R_i$ in $\mathit{BaseRank}(K)$ that does not appear at rank $i$ in $S$. Then either $\delta$ is placed at a higher rank (contradicted above), or not ranked at all (which is forbidden by the choice rule). Hence, $\delta$ must appear in $K^S_i$.

It follows that the minimal answer set must assign each defeasible conditional to the same rank as in $\mathit{BaseRank}(K)$. Therefore, ${\cal K}^{S}_{i} = {\cal R}_{i}$ for all $i$.
\end{appendixproof}

\subsubsection{Rational Closure}
We denote the Rational Closure ASP algorithm as $\mathtt{RC}$ and $\Pi$ as $\mathtt{encoding(K) \cup \mathtt{baseRank} \cup RC}$. Furthermore, for any given answer set $\cal S$ we denote $\bigcup_{j=n}^\infty \overrightarrow{{\cal K}^{\cal S}_j} =  \{$ A $\entails$ B  $ \mid \mathtt{rank(m\_implication(A,B),i)}$ $ \in \cal S\}$

We first show that our encoding is sound:

\begin{Propositionrep}
    Given some knowledge base $\mathcal{K}$ and  query A$\entails $ B, if $\mathtt{entailed(true)\in} \cal S$, where $\cal S$ is the minimal answer set, then $\mathcal{K} \mid\!\approx_{RC}$ A $\entails$ B. 
\end{Propositionrep}

\begin{appendixproof}
    Given $\mathtt{entailed(true) \in \cal S}$, by the definition of entailed in $\mathtt{RC}$ this means $\mathtt{inference(A,B)} \in \cal S$. To show that $\mathcal{K} \mid\!\approx_{RC}$ we will construct the corresponding BaseRank model using $\Pi$. Let $\bigcup_{j=n}^\infty \overrightarrow{{\cal K}^{\cal S}_j}$ = $\{ A \rightarrow B \mid \mathtt{rank(m\_implication(A,B),i) \in \cal S, \mathtt{guess(N)} \in \cal S}\}$, where $n=N$ for $\mathtt{guess(N)} \in \cal S$. By Proposition \ref{prop:baserankeq} $\bigcup_{j=n}^\infty \overrightarrow{{\cal K}^{\cal S}_j}$ maps onto a corresponding minimal base rank model used for RC, $\bigcup_{j=n}^\infty \overrightarrow{{\cal K}^{\cal S}_j}$ = $\cal R _\infty \bigcup \cal R$. The inference rules mapped out in $\mathtt{RC}$ are exactly those used for modeling entailment, thus $\cal R _\infty \bigcup \cal R \models$ A $\rightarrow B$ leading to $\mathcal{K} \mid \! \approx_{RC} A \rightarrow B$
\end{appendixproof}

Likewise, the encoding is complete:
\begin{Propositionrep}
     Given some knowledge base $\mathcal{K}$ and  query A$\entails $ B, if $\mathcal{K} \mid\!\approx_{RC}$ A $\entails$ B then there exists a minimal answer set $\cal S$ of $\Pi$ where $\mathtt{entailed(true) \in \cal S}$. 
\end{Propositionrep}

\begin{appendixproof}
$\mathcal{K} \mid\!\approx_{RC} A \entails B$, thus there exists some partition of the BaseRank model, $\cal R '$ where $\cal R_\infty \bigcup \cal R' \models$ A $\rightarrow B$. Proposition \ref{prop:baserankeq} ensures a corresponding answer set $\cal S$ which respects this ranking. Using the inference rules defined by the ASP encoding, it will then be derived that $\mathtt{inference(A,B) \in \cal S}$ with $\mathtt{query(A,B)}\in \cal S$ resulting in $\mathtt{entailed(true)} \in \cal S$. This holds as the inference rules defined follow the natural deduction style derivation rules, which are sound and complete.  
\end{appendixproof}

\section{Experiments and Results}
\label{experiments}
To assess the computational efficiency of our RC definition, we conducted various experiments testing the runtime of answering queries from a knowledge base. The knowledge bases used consisted of formulas of the form $\alpha \entails \beta$. The queries were statements explicit in the knowledge base. The following section outlines the experimental process and the tests run. We tested our implementation against the InfOCF solver's implementation of System Z.

Experiments were conducted to determine the runtime required to compute answer sets when using our ASP RC definition for entailment. Tests were conducted on a MacBook Air with an M1 processor, featuring an eight-core CPU and a seven-core GPU. 
The timing for computing entailment using our definition of RC was captured from the statistics captured by the Clasp solver once the process was complete. To compute the timing of RC within the InfOCF solver, we tested the time taken to compute inference using the time function from the Python library. It is important to note that the version of the InfOCF solver tested against is written using Python 3. 

Our experiments consist of two variables, the size of a knowledge base and the time taken to compute inference on the knowledge base. We tested knowledge bases starting from 25 statements to 325 statements in increments of 25. For each knowledge base, we tested the time taken to compute the inference of a specified query. Furthermore, we ran 2000 independent trials to represent the time taken accurately. Testing was halted beyond 325 statements due to a recursion error in the InfOCF solver.

\begin{figure}[H] \centering \includegraphics[scale=0.6]{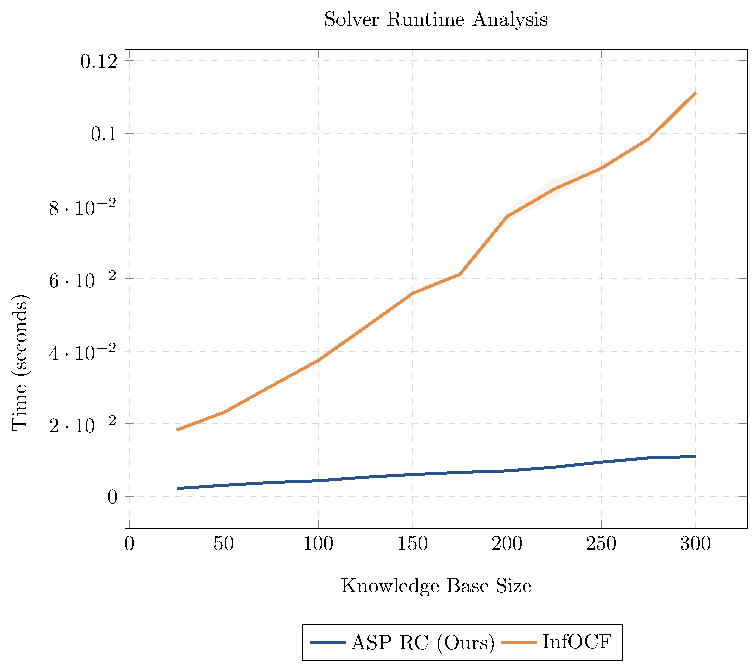} \caption{Comparison of our ASP implementation of RC and the InfOCF solver} \label{fig:experiments_results} \end{figure}

Figure \ref{fig:experiments_results} presents the average runtime (±95\% CI) comparing our ASP-based Rational Closure solver with the InfOCF solver implementation, across knowledge base sizes ranging from 25 to 325 statements. The ASP RC implementation consistently demonstrates significantly lower runtime across all KB sizes, clearly highlighting its computational efficiency. Notably, the runtime difference between the two solvers widens as the KB size increases.

The InfOCF solver shows a linear increase in runtime proportional to KB size, whereas our method's runtime scales significantly better as the size of the knowledge base increases. The least square regression of the ASP solver is 0.00003x + 0.0013, while the InfOCF solver is 0.00035x + 0.0050. Comparing the gradients of each, we notice that our method runs 12 times faster. 

The confidence intervals (CIs) further illustrate the efficiency difference: ASP’s CI remains consistent at ±1 ms even at the largest tested KB size (325), indicating a stable and predictable performance. In contrast, the InfOCF solver’s CI expands to approximately ±3 ms around 275 statements, reflecting greater sensitivity in runtime.

Quantitatively, at the largest KB size, our method achieves a runtime of approximately 0.011 seconds compared to the InfOCF solver’s 0.12 seconds, representing roughly a 91\% reduction in computational time. These findings indicate that the ASP-based RC solver offers both greater efficiency and improved scalability for entailment reasoning tasks.

Beyond computational performance, the declarative nature of ASP itself presents substantial practical advantages. ASP's clear syntax, consisting of logical implications, makes programs highly readable once the basic logical operations are understood. 

Furthermore, ASP encodings of RC allow ease of modification and extension. For instance, extending our RC solver to support Lexicographic Closure requires only adjustments within the ranking module without requiring a complete redesign of the rules of the program.
Additionally, the transparency of ASP allows for straightforward insight into the reasoning process. Since ASP rules directly represent logical specifications, it is immediately clear how answer sets are derived from given premises. This facilitates an easier debugging process and improves long-term maintainability.

\section{Conclusion and Future Work}
This paper proposed a novel declarative definition for computing entailment under RC framework using ASP. We outlined the underlying intuition for our approach and provided a formal proof of the correctness of both BaseRank and RC, defined using ASP. Additionally, we conducted experimental evaluations comparing our implementation with the InfOCF solver, an imperative alternative for computing RC. The results demonstrated that our ASP-based definition significantly outperformed the InfOCF solver in terms of computational runtime.

Beyond a decrease in computational runtime, we found that ASP has several other advantages for its use within the KLM framework. One lies in its declarative nature, which emphasises the what of a problem rather than the how. We found that this abstraction encouraged a deeper conceptual understanding of RC. In modelling RC, the declarative paradigm not only led to a clearer specification of the problem but also offered a fresh perspective on the underlying logic of KLM entailment relations. Another benefit is the compact and readable nature of the encoding.

The definition of RC presented in this paper considers works with knowledge bases consisting of statements of the form $\alpha \entails \beta$ where $\alpha$ and $\beta$ are literals. Future work could extend this by exploring more complex logical statements involving conjunctions and disjunctions. Furthermore,
while RC is one of the more well-known defeasible forms of reasoning, there are many other forms of defeasible entailment outlined by the KLM framework that can be investigated using ASP in future work. These include Lexicographic Closure \cite{lehmann_1995} and Relevant Closure \cite{relevant}.

\section*{Acknowledgements}
The authors wish to thank the anonymous reviewers for their helpful comments. 
This work is based on the research supported in part by the National Research Foundation of South Africa (REFERENCE NO: SAI240823262612). Jesse Heyninck was partially supported by the project {LogicLM}: Combining Logic Programs with Language Model with the number NGF.1609.241.010 of the research programme NGF AiNed XS Europa 2024-1, which is (partly) financed by the Dutch Research Council (NWO).
\bibliographystyle{eptcs}
\bibliography{generic}
\appendix
\section{Appendix}
\subsection{Natural deduction system}
\label{naturalDeductionSystem}
A deductive system allows for the construction of proofs and the step-by-step establishment of tautologies. The system we use in our ASP encodings is natural deduction. The natural deduction system allows for the proof of conclusions via intermediate conclusions from a certain set of premises.

The natural deduction system is both sound and complete \cite{logicmanual}. Formally, if $\Gamma$ is defined as a set of premises and $\omega$ is the sentence that is being proved, then it is possible to pass from the set of premises $\Gamma$ to $\omega$ iff $\Gamma \models \omega$. Furthermore, this system is complete in that all valid arguments $\omega$, where $\Gamma \models \omega$, are reachable by going through the proof steps defined within this system.

Proofs in natural deduction start with an assumption. To conclude that $A \rightarrow B$ holds, we will start with the assumption of $A$ and then conclude $A \rightarrow B$ from this.

In natural deduction, there are two important concepts when it comes to working with each connective: the introduction rule and the elimination rule. The introduction rule specifies how we introduce a connective to a proof, while the elimination rule specifies how we can eliminate connectives from proofs. 

\subsubsection*{Proof of Transitivity}
Given premises: $A \rightarrow B$, $B \rightarrow C$.

\[
\frac{
  \displaystyle
  \frac{\displaystyle A, A \rightarrow B}{\displaystyle B} \quad B \rightarrow C
}{
  \displaystyle
  \frac{\displaystyle C}{\displaystyle A \rightarrow C}
}
\]

Discharging the assumption, we obtain $A \rightarrow C$.

\subsubsection*{Proof of Contrapositive}
Given premise: $A \rightarrow B$.
\[
\frac{\displaystyle
    \frac{\displaystyle
        \neg B, \frac{\displaystyle A, A \rightarrow B}{\displaystyle B \quad (\rightarrow E)}
    }{\bot \quad (\neg E)}
}{\displaystyle \neg A \quad (\neg I)}
{\displaystyle \neg B \rightarrow \neg A \quad (\rightarrow I)}
\]

\subsubsection*{Proof of Disjunction Elimination}
Given premises: $A \lor \neg A$, $A \rightarrow C$, $\neg A \rightarrow C$.
\[
\frac{
    \displaystyle A \lor \neg A, \quad \frac{\displaystyle A, A \rightarrow C}{\displaystyle C }, \quad \frac{\displaystyle \neg A, \neg A \rightarrow C}{\displaystyle C }
}{\displaystyle C }
\]

\subsubsection*{Proof of Identity}
Given premise: $X \rightarrow Y$.
\[
\frac{\displaystyle X \rightarrow Y}{\displaystyle X \rightarrow Y }
\]

\end{document}